\documentclass[journal]{IEEEtran}
\ifCLASSINFOpdf
\else
\fi

\usepackage{times}
\usepackage{xcolor}
\usepackage{soul}
\usepackage[hidelinks]{hyperref}
\usepackage{graphicx}
\usepackage[caption=false]{subfig}
\usepackage{algorithm}
\usepackage{algorithmic}
\usepackage{amsmath}
\usepackage{multirow}
\usepackage{makecell}
\usepackage{url}
\usepackage{amssymb}
\begin{document}
%
\title{NAS-TC: Neural Architecture Search on Temporal \\Convolutions for Complex Action Recognition}
%
%
%

\author{Pengzhen~Ren,
        Gang~Xiao,
        Xiaojun~Chang, 
        Yun~Xiao,  
        Zhihui~Li, 
        and~Xiaojiang~Chen
\thanks{ ~Corresponding author: Yun Xiao.}
\thanks{P. Ren, Y. Xiao, and X. Chen are with School of Information Science and Technology, Northwest University. (Email: pzhren@foxmail.com, yxiao@nwu.edu.cn, xjchen@nwu.edu.cn, xcwang@ucalgary.ca)}
\thanks{G. Xiao is a senior engineer in Huawei Technologies Co.Ltd, China.zzz}
\thanks{X. Chang is with Faculty of Information Technology, Monash University, Australia, 3168. (E-mail: cxj273@gmail.com)}
\thanks{Z. Li is with Shandong Artificial Intelligence Institute, Qilu University of Technology. (E-mail: zhihuilics@gmail.com)}
}

\markboth{Journal of \LaTeX\ Class Files,~Vol.~14, No.~8, August~2015}%
{Shell \MakeLowercase{\textit{et al.}}: Bare Demo of IEEEtran.cls for IEEE Journals}
%



\maketitle

\begin{abstract}
In the field of complex action recognition in videos, the quality of the designed model plays a crucial role in the final performance. However, artificially designed network structures often rely heavily on the researchers' knowledge and experience. Accordingly, because of the automated design of its network structure, \textit{Neural architecture search} (NAS) has achieved great success in the image processing field and attracted substantial research attention in recent years. Although some NAS methods have reduced the number of GPU search days required to single digits in the image field, directly using 3D convolution to extend NAS to the video field is still likely to produce a surge in computing volume. To address this challenge, we propose a new processing framework called Neural Architecture Search-Temporal Convolutional (NAS-TC). Our proposed framework is divided into two phases. In the first phase, the classical CNN network is used as the backbone network to complete the computationally intensive feature extraction task. In the second stage, a simple stitching search to the cell is used to complete the relatively lightweight long-range temporal-dependent information extraction. This ensures our method will have more reasonable parameter assignments and can handle minute-level videos. Finally, we conduct sufficient experiments on multiple benchmark datasets and obtain competitive recognition accuracy.
\end{abstract}

\IEEEpeerreviewmaketitle

\begin{figure}
	\centering
	\includegraphics[width=1.0\linewidth]{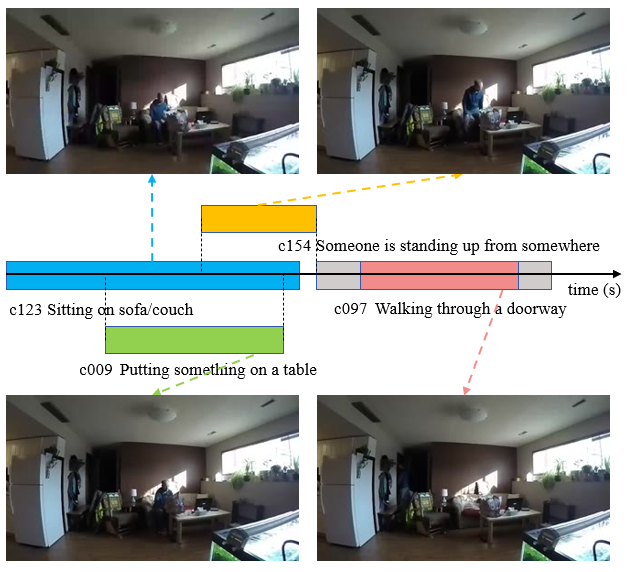}
	\caption{A complex action example CUCBN in a long video in the data set Charades \cite{Sigurdsson2016}. The video sample CUCBN has a duration of 20.88s and a total of 502 video frames, which contain 4 overlapping unit action segments (they are represented by 4 long rectangles of different colors, and the gray long rectangles are unlabeled meaningless transitional action clips).}
	\label{fig:CUCBN}
\end{figure}

\section{Introduction}
	\IEEEPARstart{D}{eep} learning has achieved unprecedented success in the field of image vision \cite{chen2020contour,chen2021bidirectional, das2019two, zhao2020learning,chang2018rcaa,Zhu0CL20}. The task of recognizing complex actions in long untrimmed videos \cite{kuehne2014language} is far closer to real-world human vision than short trimmed video recognition. The study of such tasks is consequently of great practical significance. However, applying deep learning in the video field has two obvious challenges. One is that artificially designed models are too complex and tend to rely heavily on the prior knowledge and experience of researchers. The other is that the direct use of 3D convolution to extract spatiotemporal features typically leads to a spike in computational load.

	Working in the neural architecture context, we know that the complexity of the network model in deep learning determines the model's learning potential~\cite{ChangYYX16}. More complex models require larger training sets and are thus more difficult to train; moreover, models that are too simple often fail to achieve the desired performance due to insufficient learning potential. Designing a good and moderately complex network architecture therefore requires a lot of effort from researchers, and is also difficult to judge artificially. The successful application of NAS in the image vision field has provided researchers with new possibilities to avoid the manual design of network architectures \cite{Zoph2018,RealAHL19,ren2020comprehensive, liu2020block,ChengZHDCLDG20}.

	Unfortunately, although previous NAS algorithms based on reinforcement learning (RL) \cite{Zoph2018} achieve excellent performance, the computational cost (around 2000 GPU days) is prohibitive for most researchers. The computational load in one framework search employing evolutionary algorithms reached 3150 GPU days \cite{RealAHL19}. Therefore, a large number of outstanding works have focused on how to reduce the amount of calculation and accelerate the architecture search process \cite{Cai2018a}. In addition, there are architecture search methods based on MCTS \cite{Negrinho2017}, SMBO \cite{Liu2018} or Bayesian optimization \cite{Kandasamy2018}. However, these methods treat architecture search as a black-box optimization problem on discrete domains, which makes it extremely difficult to optimize the model.
	Fortunately, a new differentiable architecture search method was proposed in Differentiable architecture search (DARTS) \cite{Liu2019}. These authors relaxed the discrete search space to make it continuous, then used gradient descent to optimize the model; in this way, the amount of calculation required by the architecture search can be greatly reduced so that ordinary researchers can feel the charm of architecture search. This research thus rapidly attracted the interest of a large number of researchers \cite{li2020sgas,zhang2020overcoming}.

    \begin{figure}
    	\centering
    	\subfloat[Search space]{
        	\includegraphics[width=0.95 \linewidth]{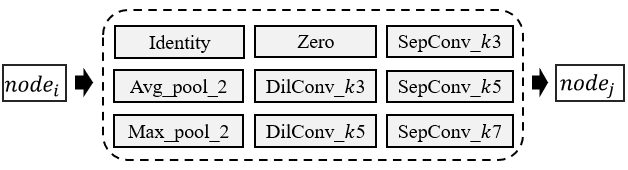}
        	\label{fig:searchspace}
    	}\\
    	
    	\subfloat[SepConv\_$k$* and DilConv\_$k$*]{
        	\includegraphics[width=0.75 \linewidth]{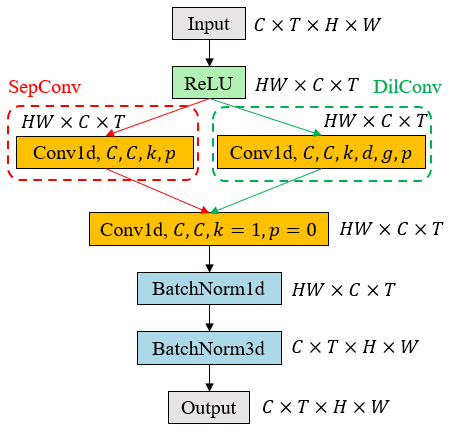}
        	\label{fig:Sep_DilConv}
    	}
    	\caption{Schematic diagram of search space and unit operations. (a) search space. Here, $ node_i $ and $ node_j $ are the $ i $-th and $ j $-th nodes in the cell. The nine operations in the middle are the search space of the NAS. The operations in the search space are all one-dimensional operations on $ \mathcal{T} $ (if applicable). (b) The execution details of the unit operations SepConv\_$k$* and DilConv\_$k$ in the search space. $C$ is the number of channels of the current feature.}
    \end{figure}
	
	This paper focuses on the recognition of complex actions in long videos. Complex actions of this kind often last for a long time and include multiple unit actions. These unit-actions not only vary greatly in terms of their temporal order, but also have uneven duration. And these unit actions overlap each other on the timeline. In Fig.\ref{fig:CUCBN}, we show a complex action example CUCBN in a long video in the data set Charades \cite{Sigurdsson2016}.
	For this reason, many innovative and complex neural architectures have been proposed and have achieved quite good performance \cite{Hussein2019,wan2020action,zhang2020knowledge,ChangNMYZ15,ZhuHCSS17,ZhangYCWCL20,YanZCLYH20,ChengGCSHZ18}. 
	However, while neural architecture design in the video field still faces similar challenges as are encountered in the image vision field, it is not possible to simply translate architecture search techniques that were successful in the image field to the video field with its complex action recognition. That is because the application of 3D convolution in the video field is more complicated than is the case for the 2D image vision task. Moreover, the image size of frames in the video field is also larger than in image vision tasks. In addition, the large amount of calculation required makes it difficult for the architecture search to complete the corresponding tasks while also using 3D convolution to extract the spatiotemporal features. To reduce the computational load, researchers began to consider decomposing the 3D convolutions. For example, CoST \cite{Li2019} proposed decomposing the 3D convolutions into three 2D convolutions; moreover, S3D \cite{Xie2018} proposed using 2+1D convolutions rather than traditional 3D convolutions, while Timeception \cite{Hussein2019} solves spatiotemporal convolutions into depthwise-separable temporal convolutions.
	
	It is worth mentioning that Timeception \cite{Hussein2019} uses CNN as the model's backbone network. This backbone network is responsible for extracting the spatiotemporal features. In this case, it completes the task of extracting the temporal features of complex actions in long videos by stacking carefully designed Timeception layers. This approach turns the original complex backbone network into a preprocessing model for spatiotemporal feature extraction. Subsequent lightweight temporal feature extraction tasks offer the possibility of using NAS.
	
	In our proposed method, we also use CNN as the backbone network. The difference is that, in our case, the design of the Timeception layer is done through NAS. This makes our model able to get rid of the limitations of manual design to a certain extent. The contributions of this paper can be briefly described as follows:
	\begin{itemize}
		\item We propose a two-phase framework specifically designed for complex action recognition in long video. This structure can allocate the parameters of the model appropriately. The amount of parameters decreased by 36\% (when TC Layers is equal to 6).
		
		\item We propose an automated architecture search method for video action recognition, which successfully extended DARTS to the video field and form a new approach, Temp-DARTS. 
		
		\item We design a special search space for video action recognition, which allows our method to better learn the temporal patterns in minute-level long videos.
		
		\item We conduct sufficient experiments on a benchmark dataset to verify the effectiveness of the proposed model. The experimental performance is comparable to that of the most advanced algorithms.
	\end{itemize}	
    
    The rest of this article is organized as follows. In Section \ref{sec: Related Works}, some related work before introducing the NAS-TC method is introduced. Then, we discuss the composition details of the NAS-TC overall framework in Section \ref{sec: Method}. In Section \ref{sec: Experiments}, we compare the performance of the proposed NAS-TC with the latest related methods on three real long video data sets. Finally, the concluding remarks are placed in Section \ref{Sec: Conclusion}.

    \begin{table}
    \begin{center}
    \caption{Search space operation information. $C$ is the number of channels of the current feature.}
    \scalebox{1}{
    	\begin{tabular}{ccccc}
    		\hline 
    		$\mathcal{O}$& \makecell[c]{kernel\_size\\($k$)} & \makecell[c]{dilation\\($d$)}& \makecell[c]{groups\\($g$)}& \makecell[c]{padding\\($p$)}\\ 
    		\hline 
    		Identity& - & - &-&- \\ 
    		Zero& - & - &-&- \\ 
    		Avg\_pool\_2& (2,1,1) & - &-&(1,0,0) \\ 
    		Max\_pool\_2& (2,1,1) & - &-&(1,0,0) \\ 
    		DilConv\_$k$3 & (3,1,1) & 2 &$C$&(2,0,0) \\ 
    		DilConv\_$k$5 & (5,1,1) & 2 &$C$&(4,0,0) \\ 
    		SepConv\_$k$3 & (3,1,1) & 1 &-&(1,0,0) \\ 
    		SepConv\_$k$5 & (5,1,1) & 1 &-&(2,0,0) \\ 
    		SepConv\_$k$7 & (7,1,1) & 1 &-&(3,0,0) \\ 
    		\hline 
    	\end{tabular}}
    	\label{tab:search_space}
    \end{center}
    \end{table}
	
    \section{Related Works}
    \label{sec: Related Works}
    \subsection{Neural Architecture Search}
    Neural architecture search is regarded as a revolution in neural network architecture design. It is expected to break through the limitations of human experience and thinking mode, so NAS is given high hopes by researchers \cite{ren2020comprehensive}. Previous NAS work based on reinforcement learning \cite{ZophL17, Zoph2018, CaiYZHY18} and evolutionary learning \cite{RealMSSSTLK17, RealAHL19} is often too time-consuming and requires a lot of computing power, so NAS cannot be widely used. 
    
    The emergence of DARTS \cite{Liu2019} has greatly improved this situation. It uses the popular cell-based \cite{Zoph2018} search space and forms the final network architecture by searching for two different types of cells (normal cell and reduction cell) and stacking them alternately. In particular, DARTS serializes candidate operations that are discretized between pairs of nodes in the cell template. Therefore, the gradient descent method is used to solve the problem of selecting candidate operations between node pairs. The continuation strategy of candidate operations between this pair of nodes can be expressed as follows:
    \begin{equation}
        x_j = \sum_{o\in \mathcal{O}}\frac{exp(\alpha^{(i,j)}_o)}{\sum_{o'\in \mathcal{O}}exp(\alpha^{(i,j)}_{o'})}x_i,
    \end{equation}
    where $\mathcal{O}$ is the operation set in Fig. \ref{fig:searchspace}, while $\alpha_o^{(i,j)}$ sents the weight of operation $o$ on directed edge $e^{(i,j)}$. The discretization of the mixing operation between node pairs can be expressed as follows:
    \begin{equation}
        o^{(i,j)} = argmax_{o\in \mathcal{O}}\alpha_o^{(i,j)}.
    \end{equation}
    Subsequently, a large number of improvements related to DARTS were proposed \cite{ChenXW019, XuX0CQ0X20, ZelaESMBH20}. 
    Although DARTS greatly reduces the time cost of cell search, and has achieved success in the field of image vision. However, it is still difficult to directly extend DARTS to the video field.
    
    \subsection{Complex Action Recognition}
    The biggest difference between image vision task and video tasks is the information in the temporal dimension. Therefore, video tasks are more complicated than image vision tasks. Video tasks require additional modeling of temporal dimensions. Commonly used modeling methods include: statistical pooling \cite{HusseinGS17, HabibianMS17, GirdharR17, FernandoGMGT17}, vector aggregation \cite{OneataVS13, DutaIAS17, DutaIAS17}, and neural network-based methods \cite{DonahueHGRVDS15, GhodratiGS18, HuangLMW17}. Statistical pooling and vector aggregation ignore the time cues in the video task, and the above-mentioned neural network-based model is also extremely limited for information mining in the temporal dimension.
    
    Subsequently, some short-range action recognition models based on neural networks were proposed. For example, the Two-stream convolutional network \cite{SimonyanZ14} uses OpticalFlow as auxiliary information for RGB images. \cite{BilenFGV18} uses dynamic graphs for action recognition. C3D \cite{JiXYY13, TranBFTP15} and I3D \cite{CarreiraZ17} implement short-range temporal information learning by expanding 2D convolution. These methods are often computationally expensive and cannot effectively learn temporal information in long video clips. 
    
    To enable the model to learn the complex temporal patterns in long videos, TRN \cite{zhou2018temporal} studies the temporal dependence between different video frames on multiple temporal scales; TSN \cite{WangXW0LTG16, 0002X00LTG19} divides the long video into video segments, and integrates the temporal patterns of different video segments to make video-level predictions; Non-local networks \cite{Wang2018} introduces the attention mechanism to the learning of temporal patterns of long videos. There are many researches related to long video complex action recognition, such as \cite{sigurdsson2017asynchronous, VarolLS18}. However, the above method can successfully model 128 timesteps at max \cite{Hussein2019}. Timeception \cite{Hussein2019} successfully modeled 1024 timesteps by carefully designing a Temporal Conv Module for the temporal dimension. However, the manually designed model is most likely not optimal. The design of an automated Temporal Conv Module is a problem worthy of study.
    
	\section{Method}
	\label{sec: Method}
	\subsection{Motivation}
	\subsubsection{The Predicament of NAS in the Video Field} \leavevmode\\
	Although DARTS \cite{Liu2019} was able to reduce the model search cost from thousands of GPU days to single digits, it is still very difficult to require DARTS to learn spatiotemporal information directly from video data using 3D convolution. In addition to the difficulty of coping with the surge in computational volume, it is also difficult to fully learn the temporal information in long videos.

	\subsubsection{Individual Convolutions on Temporal Information} \leavevmode\\
	For video information, the 3D CNN learns the spatiotemporal kernel through three orthogonal subspaces, specifically the temporal ($ \mathcal{T} $), spatial ($ \mathcal{S} $) and semantic channel subspace ($ \mathcal{C} $) \cite{Hussein2019}. In previous models, 3D convolution was often considered to convolve temporal and spatial information simultaneously \cite{Carreira2017}), and has also been considered for decomposing 3D convolutions into 2+1D convolutions \cite{Xie2018}. However, the time information processing ability of these methods is insufficient, leading to these methods being unable to handle minute-level long videos. The success of Timeception \cite{Hussein2019} illustrates to some extent the effectiveness of the convolution of temporal information alone. This approach can also alleviate the problem of the surge in computation to some extent.
	
	\subsubsection{A Simple and Effective Implementation} \leavevmode\\
	An intuitive and effective idea would be to extract the spatiotemporal features using the classic CNN as the backbone network of the model. The NAS method is then used to complete the model design of lightweight tasks for long video temporal information extraction. In this way, the original heavy video task can be broken down into two parts: the computationally expensive spatiotemporal feature extraction task becomes data preprocessing, while the lightweight design of the long-term feature extraction task is left to NAS.

    \begin{figure}
    	\centering
    	\includegraphics[width=1\linewidth]{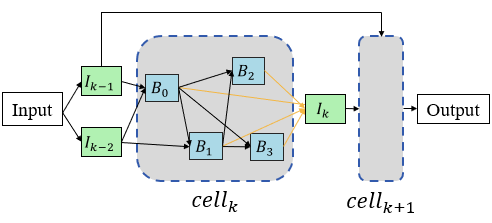}
    	\caption{A network structure diagram with multiple stacked cells. Each cell has two inputs and one output. The $ cell_{k} $ has two input nodes $ I_{k-1} $ and $ I_{k-2} $, which are the outputs of the two previous cells. $ I_k $ is the output of the $ k $-th cell, $ cell_k $. There are four intermediate nodes ($ B_0, B_1, B_2 $ and $ B_3 $) inside each cell, which are used to perform feature extraction and processing tasks. Finally, the features of the four intermediate nodes are concatenated to obtain the output $ I_k $ of $ cell_{k} $.}
    	\label{fig:cell}
    \end{figure}

	\subsection{NAS-TC: Neural Architecture Search Temporal Convolutional}
	In this section, we will briefly introduce the basic components of NAS, then focus on the construction of NAS-TC. Some settings in our model are similar to previous NAS work \cite{Liu2019}.
	
	\subsubsection{The Basic Components of NAS-TC}\leavevmode\\
	\textbf{Search Space.} As illustrated in Fig. \ref{fig:searchspace}, $ node_i $ and $ node_j $ are the $ i $-th and $ j $-th nodes in the cell. These nodes represent a latent representation, such as a feature map in a convolutional network. The nine operations in the middle are the search space $\mathcal{O}$ of the NAS. All operations in the search space are one-dimensional operations on $ \mathcal{T} $ (if applicable). The search space $\mathcal{O}$ contains \(3\times 1\times 1\), \(5\times 1\times 1\) and \(7\times 1 \times 1\) separable convolutions, \(3\times 1\times 1\) and \(5\times 1\times 1\) dilated separable convolutions, $ 2\times1\times1 $ max pooling, $ 2\times1\times1 $ average pooling, identity, and $ zero $. All operations have $ stride=1 $ (if applicable). Moreover, proper padding is used to keep the feature map in its original dimension for the feature map stitching. The order of the convolution operations used here is ReLU-Conv-BN, and the specific schematic diagram is shown in Fig \ref{fig:Sep_DilConv}.
	Information on the operation of the search space is summarized in Table \ref{tab:search_space}.

    \textbf{Cell Structure.} The cell settings in our model are consistent with those in DARTS \cite{Liu2019}. Fig. \ref{fig:cell} presents a network structure diagram using multiple stacked cells. The cell needs to be obtained using the NAS search. Each cell has two inputs and one output. The $ cell_{k} $ has two input nodes, $ I_{k-1} $ and $ I_{k-2} $, which are the outputs of the two previous cells. $ I_k $ is the output of the $ k $-th cell, $ cell_k $. There are also four intermediate nodes ($ B_0, B_1, B_2 $ and $ B_3 $) inside each cell, which are used to perform feature extraction and processing tasks. Finally, the features of the four intermediate nodes are concatenated to obtain the output $ I_k $ of the $ cell_{k} $.
	
	The directed edges $ (i,j) $ between every two nodes $ B_i $, $ B_j $ represent the operations $ o_{i,j} $. During the training phase, each intermediate node is calculated from all previous nodes as follows: \(B_j = \sum_{i<j}o_{i,j}\times B_i \). The original discretization operation is relaxed into a continuous state, after which the gradient descent method is used to learn the structure of the network model cell. During the testing phase, each node retains only the two operations with the highest weight among the associated operations. This can reduce the total amount of calculation, as well as the requirements for GPU memory to a certain extent.

	\subsubsection{Temp-DARTS} 
\begin{figure*}
    	\centering
    	\subfloat[The overall framework of NAS-TC network.]{
    	\includegraphics[width=0.4 \linewidth]{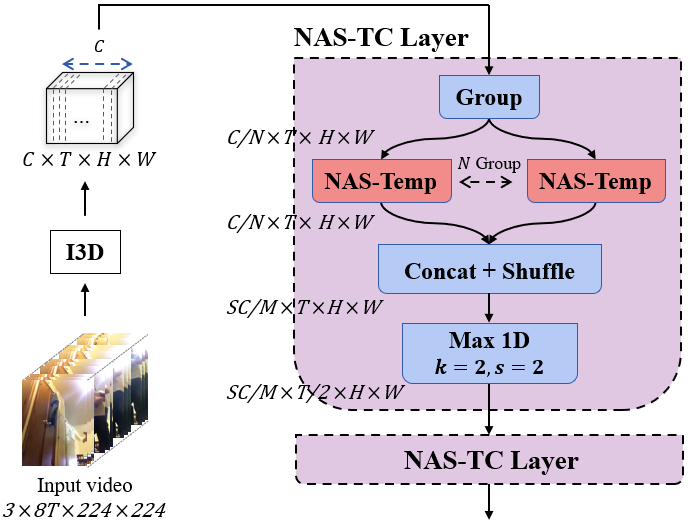}
    	\label{fig:modelarchitecture}}
    	\hspace{2 em}
    	\subfloat[NAS-Temp]{
    	\includegraphics[width=0.33 \linewidth]{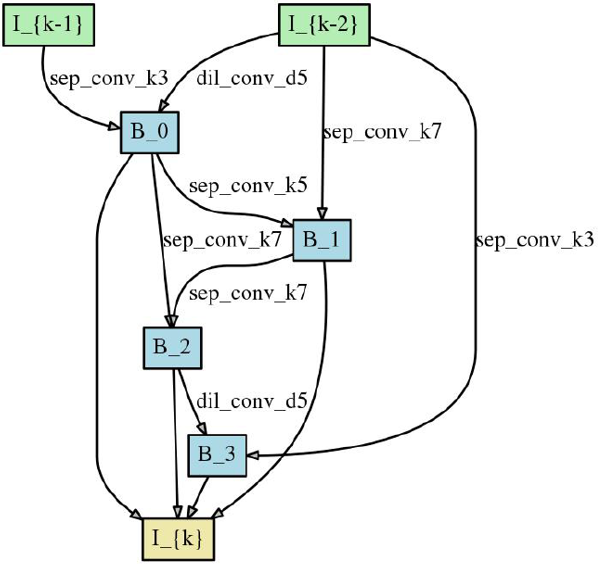}
    	\label{fig:charades_cell}}
    	\caption{NAS-TC network and NAS-Temp. (a) We use I3D as the backbone network for the model; accordingly, feature extraction is performed on the sampled video using I3D. Multiple NAS-TC layers are then simply stacked and connected to the classifier to complete the final classification task. (b) Cell learned on Charades: NAS-Temp. We let $ T = 32 $, sample on Charades, and extract features using I3D. The extracted features are then used as the input of Temp-DARTS, which enables us to find the best cell structure.}
    \end{figure*}

\leavevmode\\
    In our method, we extend DARTS \cite{Liu2019} to the temporal subspace. Since this method mainly involves searching for suitable cells in the temporal subspace, we refer to it as Temp-DARTS. In order to enable Temp-DARTS to be applied efficiently and conveniently in the video field, we have specially designed a new search space for it. The search space, $\mathcal {O} $, is shown in Fig. \ref {fig:searchspace}, while the inflated details for basic operations in the search space are shown in Table \ref{tab:search_space}. Since the searched cells perform corresponding operations only in the temporal dimension, we refer to the searched cells as NAS-Temp. The nodes and edges in NAS-Temp are consistent with the corresponding meanings in Figures \ref{fig:searchspace} and \ref{fig:cell}. Moreover, Fig. \ref{fig:charades_cell} shows the learned cell structure on the operation set $\mathcal{O}$ using Temp-DARTS on the Charades \cite{Sigurdsson2016} dataset.
	
	It should be noted that the DARTS search on the two-dimensional image begins from the most original image and needs to search for two cell structures: a normal cell and a reduction cell. Normal cells are primarily used to extract high-level features, while reduction cells are used to reduce the feature map size. Unlike two-dimensional DARTS, however, Temp-DARTS only need to search for a normal cell. This is mainly because the spatial dimension of the feature map obtained following processing by the backbone network has been reduced to a comparatively small level; thus, the spatial dimension of the feature map on the subsequent NAS-TC layer is kept unchanged.

	\subsubsection{Overall Framework} \leavevmode\\
	The overall framework of our model is illustrated in Fig. \ref{fig:modelarchitecture}. The overall framework of NAS-TC is composed of a backbone network and multiple NAS-TC layers stacked.
	
	\textbf{Backbone Network.} For a video sample of length $ t $, we first divide $ t $ into $ T $ segments, then take eight consecutive frames as a superframe for each segment. We then input the extracted $ T $ superframes together into the backbone network I3D \cite{Carreira2017} for feature extraction. We use the feature map of mixed-5c, which is the last spatiotemporal convolution layer of I3D, as the input to the subsequent network. That is, the original video sample $ 3 \times 8T \times 224 \times 224 $ uses I3D to perform feature extraction in order to obtain the feature map of $ C \times T\times H\times W $; here, $C = 1024, H = W = 7$. 
	
	\textbf{NAS-TC Layer.} On the dataset Charades, we let $ T = 32 $, and then extract features through I3D. We use the operation in Fig. \ref{fig:searchspace} as the search space for Temp-DARTS, while the extracted features become the input for Temp-DARTS. We next search the structure of the cell on the Charades training set. The structure of the searched cell NAS-Temp is illustrated in Fig. \ref{fig:charades_cell}. It should be pointed out that although the operations used in the NAS-Temp searched in Fig. \ref{fig:charades_cell} do not contain all single operations in the operation set, this does not mean that single operations not appearing in the search space can be directly removed from the operation set. After repeating the cell architecture search experiment under different initializations, we found that the final cells contained different unit operations, but had similar performance. This means that different unit operation combinations in the operation set may have similar learning capabilities; that is, there may be more than one optimal network structure, which is in line with what common sense suggests. Therefore, these unit operations are indispensable in cell structure search research.

	In the NAS-TC layer, the feature map $X$ extracted by I3D is divided into $ N $ channel groups. Each group feature $ x_i\in R^{\frac{C}{N}\times T \times H \times W} $ is then used as the input of NAS-Temp. To reduce the number of parameters, the channel shuffle operation \cite{Zhang2018} is then performed on the obtained features of the $ N $ NAS-Temps, enabling the learning of information across the channels. $ 2\times 1\times 1 $ max pooling is performed on $ \mathcal{T} $ at the end of the NAS-TC layer. The dimension of the obtained feature is $ \frac{SC}{M} \times \frac{T}{2} \times H \times W$. Here, $ S = 4 $, which is the number of intermediate nodes in the NAS-Temp cell, while $M =3; \frac{S}{M}$ is the scaling ratio of the channel. This is in accordance with the design principle of incremental information extraction and corresponding feature dimension halving \cite{Simonyan2014}. Subsequently, multiple NAS-TC layers are simply stacked and connected to the classifier, enabling the final classification task to be completed.

	\subsubsection{Implementation Details} \leavevmode\\
	Cells are searched on Charades' training set by Temp-DARTS, with the number of epochs set to 50. When using the backbone network for feature extraction, we apply the same settings as in \cite{Hussein2019} to perform pre-training on a specific dataset, which allows us to obtain high-quality features. The pre-training weights are publicly available\footnote{\textit{https://github.com/piergiaj/pytorch-i3d/blob/master/models/\\rgb\_charades.pt} and 
	\textit{https://github.com/hassony2/kinetics\_i3d\_\\pytorch/blob/master/model/model\_rgb.pth}}. After obtaining the corresponding features, we insert multiple NAS-TC and MLP layers on top of the backbone network to achieve the final classification. Throughout the entire training process, only the parameters inserted in this part are trained. The model is trained with a batch size of 18 for 300 epoch. We here use the Adam optimizer, where $ lr = 0.01 $ and $ eps = 1e-4 $.
	
    \begin{figure*}[!tp] 
    	\centering 
    	\subfloat[Temporal Conv Module.] 
     		{
     			\includegraphics[width=0.38\textwidth]{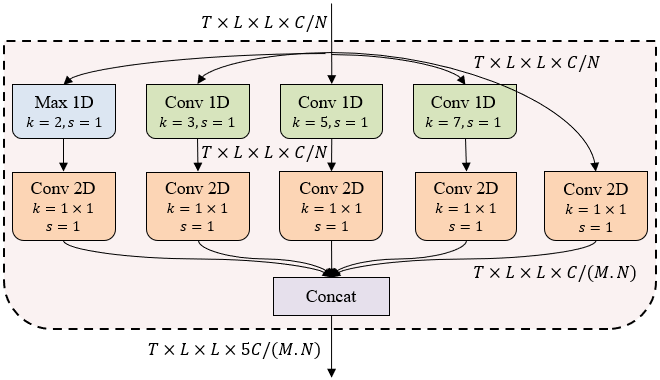} 
     			\label{fig:b_Temporal_Conv_Module}
     		}
     		\hspace{0.2em}
     \subfloat[Expanded view of NAS-Temp in Fig. \ref{fig:charades_cell}.] 
     		{
     			\includegraphics[width=0.58\textwidth]{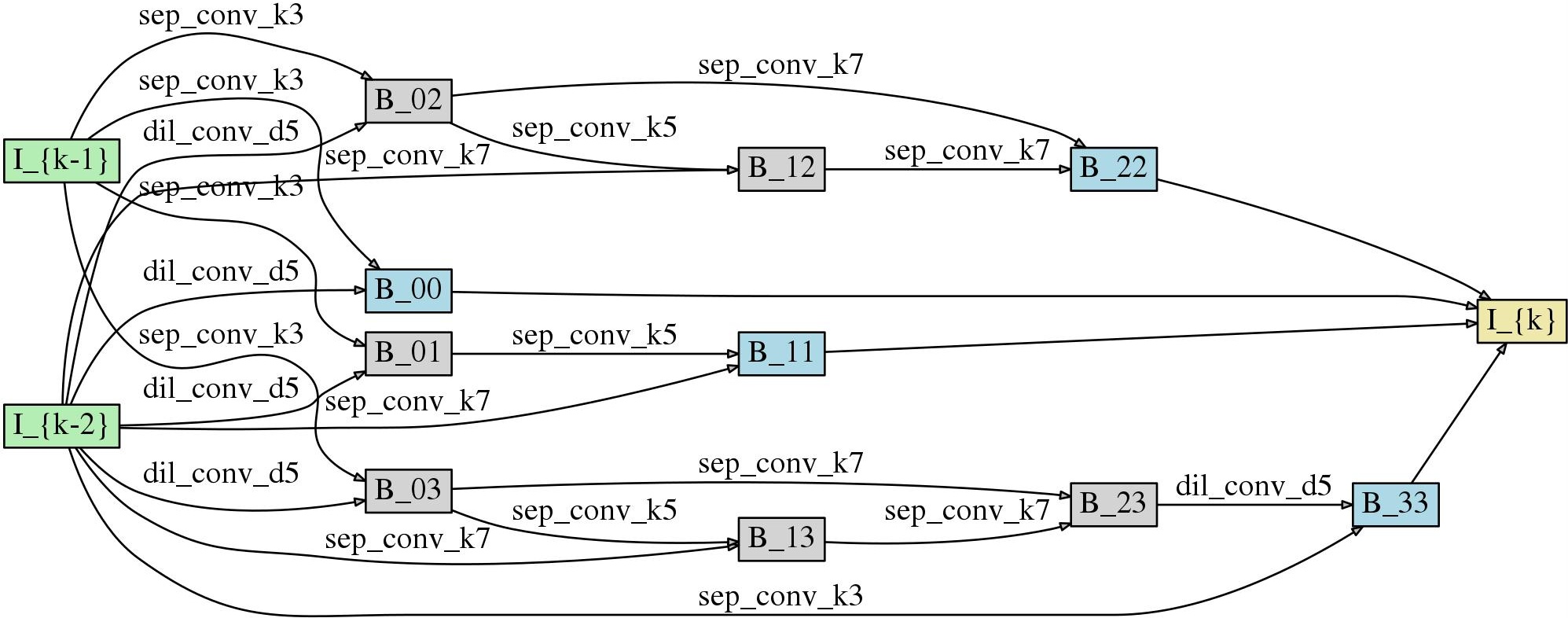}
     			\label{fig:a_NAS_TEmp}
     		}
     		\caption{Comparison of NAS-Temp and Temporal Conv Module. (a) The Temporal Conv Module is a multi-scale convolution operation similar to Inception in Timeception; the difference is that the Temporal Conv Module only performs convolution on the temporal dimension. The Temporal Conv Module was reproduced from (Hussein et al. 2019). (b) An expanded view of the four intermediate nodes of NAS-Temp in Fig. \ref{fig:charades_cell}. The corresponding nodes of the four intermediate nodes ($ B_0, B_1, B_2 $ and $ B_3 $) in the expanded view are ($ B_{00}, B_{11}, B_{22} $ and $ B_{33} $). For $ B_{ij} $, $ i $ represents the $ i $-th intermediate node in NAS-Temp, while $ j $ represents that the node is associated with the generated $ j $-th node. The input and output nodes $ I_{k-2}, I_{k-1} $ and $ I_{k} $ remain the same as in Fig. \ref{fig:charades_cell}.} 
    	\label{fig:Parameter and Performance Analysis} 
    \end{figure*}

	\subsection{Parameter Analysis}
	\label{sec:Parameter Analysis}
	In this section, we will focus on the rationality of the parameter allocation of NAS-TC and Timeception.
	
	We present an expanded view of the NAS-Temp and the structure of the Temporal Conv Module in Fig. \ref{fig:Parameter and Performance Analysis}. More specifically, Fig. \ref{fig:a_NAS_TEmp} is an expanded view of the four intermediate nodes generated in NAS-Temp in Fig. \ref{fig:charades_cell}. The nodes corresponding to the four intermediate nodes ($ B_0, B_1, B_2 $ and $ B_3 $) in the expanded view are ($ B_{00}, B_{11}, B_{22} $ and $ B_{33} $). For node $ B_{ij} $, $ i $ represents the $ i $-th intermediate node in NAS-Temp, while $ j $ represents that the node is associated with the generated $ j $-th node. The input nodes $ I_{k-2}$ and $I_{k-1} $ and the output node $ I_{k} $ remain the same as in Fig. \ref{fig:charades_cell}. Fig. \ref{fig:b_Temporal_Conv_Module} illustrates a core component of Timeception, namely the Temporal Conv Module. This module is a multi-scale convolution operation similar to Inception \cite{szegedy2015going} in Timeception. The difference is that the Temporal Conv Module only performs convolution on the temporal dimension. The Temporal Conv Module is reproduced from \cite{Hussein2019}. As shown in Fig. \ref{fig:b_Temporal_Conv_Module}, this module has five branches. More specifically, the fifth branch only performs a dimensionality reduction operation on the input using $ 1 \times 1 $ 2D convolution in the spatial dimension, while the remaining four branches perform either multi-scale convolution operations or pooling operations on the temporal dimension. In order to facilitate the management of the output dimension, the input is reduced by $M$ times using a $ 1 \times 1 $ 2D convolution. After the channels are reduced, the outputs of the five branches are concatenated across the channel dimensions as the output of the final Temporal Conv Module. For more details, please see \cite{Hussein2019}.
	
	In the Temporal Conv Module, although the $ 1 \times 1 $ 2D convolution seems to occupy very little in the way of a parameter budget, this is not actually the case. In fact, the $ 1 \times 1 $ 2D convolution requires a large number of parameters in each branch. More specifically, in the Temporal Conv Module, the $ 1 \times 1 $ 2D convolution takes up 57\% of the parameters in the five branches. These parameters are mainly used to reduce the number of channels of the feature map. At the same time, these convolution operations are also used to increase the model complexity, as this improves its learning potential. However, in fact, only the convolution operations ($ kernel\_size = 3, 5, 7 $) on the temporal dimension in three of the branches are actually used to learn advanced features on the temporal dimension. Therefore, after the above analysis, we have reason to be concerned about the rationality of this parameter allocation.

	By contrast, in NAS-Temp, we perform the necessary $ 1 \times 1 $ 2D convolution on only two inputs, specifically, $ I_{k-1} $ and $ I_{k-2} $. The number of channels of the two inputs $I_{k-1}$ and $ I_{k-2} $ is compressed to one-third of $ I_{k-1} $ by means of a $ 1 \times 1 $ 2D convolution. In the four intermediate nodes (\(B_0, B_1, B_2, B_3\)), the number of channels is kept unchanged. From the expanded view of NAS-Temp in Fig. \ref{fig:a_NAS_TEmp}, we can clearly see how these intermediate nodes are reused. However, this generation method makes the generation of each intermediate node very complicated. For example, the number of unit operations performed in order to generate $ B_{00}, B_{11}, B_{22} $ and $ B_{33} $ is 2, 4, 6 and 8, respectively. Compared with the Temporal Conv Module, this connection method makes the $ 1 \times 1 $ 2D convolution in the entire NAS-Temp layer significantly reduce the proportion of parameters. Moreover, because NAS-Temp stacks and stitches different convolution operations, the information it learns on the temporal dimension is also more complicated. This strategy gives NAS-Temp the possibility to learn more from a temporal dimension. This is in line with the principle of subspace efficiency \cite{Hussein2019}, which involves allocating a higher parameter budget to the main difficulty of the current task; that is, learning information on \(\mathcal{T}\).
	
	Therefore, our model is better at allocating the limited parameters to learning information on \(\mathcal{T}\), thereby achieving better performance. In addition, this effectively reduces the number of parameters in the NAS-Temp layer. We provide a more detailed parameter comparison in the Parameters and Search Time section.

 	\begin{table*}[htbp]
 	\begin{center}
 	\caption{Details of the datasets Charades, MultiTHUMOS and BreakfastActions datasets.}
 		\begin{tabular}{lcccccc}
 			\hline 
 			Datasets& Total videos & \makecell[c]{Total\\time (hrs)} & Average duration & \makecell[c]{Actions\\per video} & Classes & Type \\ 
 			\hline 

 		    \makecell[l]{\cite{Sigurdsson2016} Charades }& 9848 & 82 &30 sec & 6 & 157 & Daily activities \\ 
 			\makecell[l]{\cite{Yeung2018} MultiTHUMOS}& 400 & 30 &4.5 min &11 & 65 &Sports\\ 
 			\makecell[l]{\cite{kuehne2014language} BreakfastActions} & 1712 & 66 &2.3 min & 7 & 48 & Cooking \\ 	
 			\hline
 		\end{tabular}
 		\label{tab:datasets}
 	\end{center}
    \end{table*}

	\section{Experiments}
	\label{sec: Experiments}
	\subsection{Datasets}
	In order to verify the model's recognition performance on complex actions in long videos, we choose to test the performance of the model on the datasets Charades\footnote{\textit{https://prior.allenai.org/projects/charades}} \cite{Sigurdsson2016}, MultiTHUMOS\footnote{\textit{http://ai.stanford.edu/\(\sim\)syyeung/everymoment.html}} \cite{Yeung2018} and BreakfastAction\footnote{\textit{https://serre-lab.clps.brown.edu/resource/breakfast-actions-dataset/}} \cite{kuehne2014language} from the perspective of the video action composition, the temporal extent and the temporal order. Those datasets whose video length is less than half a minute or even shorter are not included in the scope of this paper. The details of the Charades, MultiTHUMOS and BreakfastActions datasets are summarized in Table \ref{tab:datasets}.
	
	\textbf{Charades} is a large-scale multi-label video classification dataset. It contains a large number of human daily activities. This dataset contains 9848 videos with a total of 157 categories, including an 8k training set (67 hrs) and a 1.8k validation set. Each video is up to 30 seconds in length and contains 6.8 unit-actions, which meets the requirements of long video complex action recognition tasks. It is therefore a very challenging benchmark dataset. For comparison, we use the same evaluation index as \cite{Wang2018,Hussein2019}, namely mean Average Precision (mAP) \cite{Pedregosa2011}.
	
	\textbf{MultiTHUMOS} is extended from the original THUMOS-14 dataset \cite{Idrees2017}. It is also an untrimmed human activity multi-class dataset, containing 400 videos with 65 action classes for a total of 30 hours. It is often used for temporal positioning in previous work; the difference is that we use it to complete the complex action recognition task of a long video. Compared to the Charades dataset, the actions in each video of this dataset are more complicated; on average, each video contains 11 unit-actions. As with Charades, we also use mAP as the evaluation index for MultiTHUMOS.
	
	\textbf{BreakfastActions} is by far one of the largest fully annotated long video action datasets available. This dataset contains 10 actions related to breakfast preparation, recorded by 52 people in 18 different kitchens. It contains 1712 videos in total; the size of the training set and the test set are 1357 and 335 respectively, while the average video length is 2.3 minutes. Each video represents an activity consisting of a series of unit-actions. There are a total of 48 types of unit-actions. Similar to \cite{Hussein2019}, we use single-label activity annotations to calculate accuracy (Acc), and multi-label unit-action annotation to calculate mAP.

    \begin{table}[htbp]
	\centering
	\caption{Comparison results of Timeception and NAS-TC under different timesteps inputs. We use I3D as a common backbone network for Timeception and NAS-TC. We insert multiple layers of TC (Timeception or NAS-TC layers) on top of I3D to form their respective networks. \textit{CNN\_Steps}: input timesteps to the CNN; \textit{TC\_Steps}: input timesteps to the first TC layer; \textit{Params}: number of parameters used by TC layers, in millions.}

	\begin{tabular}{c|c|cc|cc}
		\hline
		\multirow{2}{*}{} & \multirow{2}{*}{\makecell{CNN\_Steps\\$\Rightarrow$ TC\_Steps}} & \multicolumn{2}{c|}{Timeception} & \multicolumn{2}{c}{NAS-TC} \\
	     &            & Params &  mAP(\%)  & Params &   mAP(\%)   \\
		\hline
		 + 3 TC  &   256 $\Rightarrow$ 32     & 2.0 &  33.9   & \textbf{1.5} &  \textbf{34.5}  \\
		  + 3 TC  &   512 $\Rightarrow$ 64      & 2.0 &  35.5   & \textbf{1.5} &  \textbf{37.2}  \\
		  + 4 TC  &   1024 $\Rightarrow$ 128     & 2.8 &  37.2   & \textbf{2.0} &  \textbf{39.3}     \\
		\hline
	\end{tabular}
	
	\label{tab:Charades_Timeception_NAS}
	\vspace{-0.2cm}
    \end{table}

	\subsection{Experiments on Charades}
	\label{sec: Experiments_on_Charades}
	\subsubsection{Comparison of Long-range Temporal Dependencies Learning Ability} \leavevmode\\
	A performance comparison is conducted of Timeception and NAS-TC under different timestep inputs. By changing the number of timesteps inputs, we can observe the model's learning ability on long-range temporal dependencies. We use I3D as a common backbone network for Timeception and NAS-TC. We insert multiple layers of TC (i.e., Timeception or NAS-TC layers) on top of I3D to form their respective networks. \textit{CNN\_Steps} is the input timesteps of CNN (i.e., I3D), while \textit{TC\_Steps} is the input timesteps of the first TC layer; moreover, \textit{Params} is the number of parameters used by TC layers (in millions). The timesteps input into the backbone network is \(\textit{CNN\_Steps} \in \{256, 512, 1024\}\); accordingly, the timesteps input to the first TC layer is \(\textit{TC\_Steps} \in \{32, 64, 128\}\).
	
    From Table \ref{tab:Charades_Timeception_NAS}, it can be seen that when \textit{CNN\_Steps} gradually increases, the performance of Timeception and NAS-TC increases monotonically. This reveals that our NAS-TC and Timeception methods have been able to learn long-range temporal information to a certain extent. However, in contrast, our method uses fewer parameters and achieves better recognition performance. This performance gain also becomes more obvious with the increase of \textit{CNN\_Steps}. This is mainly because the NAS-Temp structure learned through Temp-DARTS is able to handle long-range temporal information more effectively.

	\subsubsection{Performance Comparison} \leavevmode\\
	To test the validity of our model, in this section, we compare the performance of NAT-TC with a number of related state-of-the-art methods. These methods are as follows: Two-stream \cite{sigurdsson2017asynchronous}, Two-stream + LSTM \cite{sigurdsson2017asynchronous}, 
	ActionVLAD \cite{girdhar2017actionvlad}, 
	Temporal Fields \cite{sigurdsson2017asynchronous}, 
	Temporal Relations \cite{zhou2018temporal}, 
	ResNet-152 \cite{sigurdsson2017asynchronous}, 
	ResNet-152 + Timeception \cite{Hussein2019}, 3D ResNet-101 \cite{Wang2018}, I3D \cite{Carreira2017}, I3D + ActionVLAD \cite{girdhar2017actionvlad}, I3D + Timeception \cite{Hussein2019}, I3D-NL \cite{Wang2018}, EvaNet \cite{piergiovanni2019evolving} and I3D + VideoGraph \cite{hussein2019videograph}. The results of this performance comparison are shown in Table \ref{tab:Charades}. The experimental results reveal that our method can more effectively use the features learned by the original backbone network I3D. Moreover, compared with the artificially designed I3D + Timeception network (which is the most structurally similar), our model obtains more competitive recognition accuracy.
	
    \begin{table}
    	\begin{center}
    	\caption{Performance comparison of different methods on the Charades dataset. Compared with other methods that use the same input (that is, the backbone network is I3D), I3D + NAS-TC has obvious advantages.}
    		\begin{tabular}{lcc}
    			\hline
    			Method    & Modality & mAP(\%) \\
    			\hline
    			\cite{sigurdsson2017asynchronous} Two-stream   & RGB + Flow & 18.6 \\
    			\cite{sigurdsson2017asynchronous} Two-stream + LSTM & RGB + Flow & 17.8 \\
    			\cite{girdhar2017actionvlad} ActionVLAD   & RGB + iDT & 21.0 \\
    			\cite{sigurdsson2017asynchronous} Temporal Fields  & RGB + Flow & 22.4 \\
    			\cite{zhou2018temporal} Temporal Relations & RGB  & 25.2 \\
    			\cite{sigurdsson2017asynchronous} ResNet-152   & RGB  & 22.8 \\
    			\cite{Hussein2019} ResNet-152 + Timeception  & RGB  & 31.6 \\
    			\cite{Wang2018} 3D ResNet-101   & RGB  & 35.5 \\
    			\cite{Carreira2017} I3D    & RGB  & 32.9 \\	
    			\cite{girdhar2017actionvlad} I3D + ActionVLAD  & RGB & 35.4\\
    			\cite{Hussein2019} I3D + Timeception   & RGB  & 37.2 \\
    			\cite{Wang2018} I3D-NL   & RGB  & 37.5 \\
    			\cite{hussein2019videograph} I3D + VideoGraph   & RGB  & 37.8 \\
    			\cite{piergiovanni2019evolving} EvaNet & RGB & 38.1 \\
    			\hline
    			I3D + NAS-TC  & RGB  & \textbf{39.3}  \\
    			\hline
    		\end{tabular}
    		
    		\label{tab:Charades} 
    	\end{center}
    	\vspace{-0.8cm}
    \end{table}

	\subsubsection{Parameters and Search Time} \leavevmode\\
	\label{sec:Parameter Comparison}
	Because the architecture of Timeception \cite{Hussein2019} is the most similar to that of our model NAS-TC, in this section, we will mainly compare the changes in parameters between them as the TC layer increases. Secondly, we also report the search time of the NAS-Temp cell.
	
\begin{figure}
    	\centering
    	\includegraphics[width=0.8\linewidth]{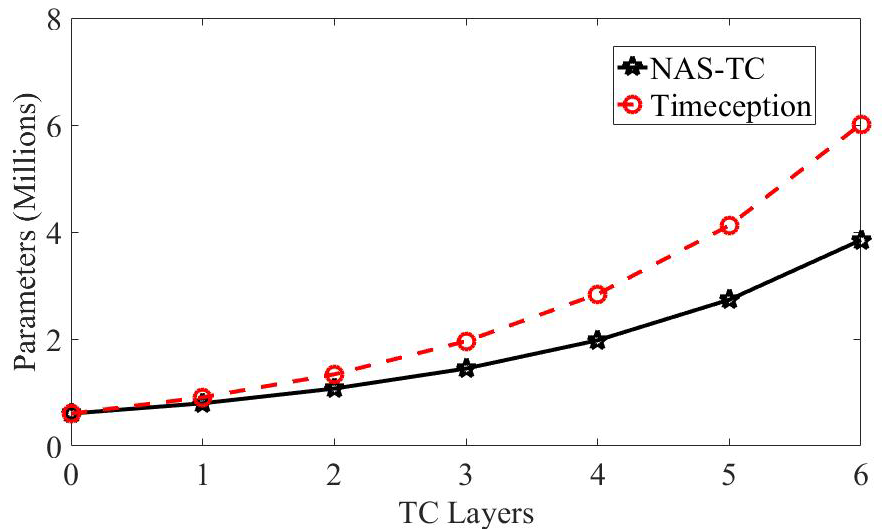}
    	\caption{As the number of TC layers changes, the parameters of NAS-TC and Timeception are compared. Obviously, with the same number of TC layers, NAS-TC has fewer parameters.}
    	\label{fig:params}
    	\vspace{-0.1cm}
    \end{figure}

	We plot the changes in the parameters of NAS-TC and Timeception with the increase of TC layer in Fig. \ref{fig:params}. When \(TC\_Layers = 0\), the model has only the top fully connected layer, and the number of parameters is 0.61 million. Clearly, NAS-TC has significantly fewer parameters than Timeception; as the number of TC layers increases, this advantage becomes more obvious. We analyzed this in detail in the Parameter Analysis section. In terms of performance, our model also exhibits obvious advantages, as can be seen from the experimental results in Table \ref{tab:Charades_Timeception_NAS}. In theory, the TC layer can also be stacked with more layers. When \(TC\_Layers = 8\), NAS-TC can process 4096 timesteps on the temporal dimension; this means that NAS-TC can handle longer videos of up to 2.3 minutes in length. Therefore, the advantages of NAS-TC (i.e. the need for fewer parameters) will be further manifested. It should be pointed out here that this does not mean that a larger number of TC layers is always better. A large TC number means a larger frame sample; since the video length is fixed, overlarge frame sampling may cause serious information redundancy problems. To facilitate comparison, including the next two datasets, we keep the number of TC layers the same for Timeception.
	
	We use Temp-DARTS to search for cells on Charades' training set. The 50 epochs take only 2.1 GPU days, which is similar to the search time achieved by DARTS in the image field. This demonstrates that it is feasible to reduce the amount of calculation by performing convolution operations only in the temporal dimension.

	\subsection{Experiments on MultiTHUMOS}
	To further verify the scalability of our model, we also test its performance on the MultiTHUMOS dataset. We use a processing method similar to that used on the Charades dataset , and compare the performance of the original I3D, Timeception, and other related models with our model. The training method is kept as consistent as possible with that of Timeception. The final experimental results are reported in Table \ref{tab:MultiTHUMOS}. From the experimental results, it is clear that multi-scale convolution is significantly better than a fixed single scale; because it is more flexible, it can more easily capture the changing temporal information in complex actions. Moreover, the I3D + Timeception method is also a multi-scale convolution operation; by comparison, because our method automatically searches for cells through NAS, our model has more flexible temporal information capture capabilities. Our model is therefore able to obtain relatively good recognition accuracy.
	
    \begin{table}[ht]
    	\begin{center}
    	\caption{Performance comparison of different methods on the MultiTHUMOS dataset. Clearly, multi-scale convolution is significantly better than a fixed single scale. NAS-TC and Timeception both use multi-scale kernels; of these two methods, NAS-TC has better accuracy.}
    		\begin{tabular}{lccc}
    			\hline
    			Method   & \makecell[c]{Kernel \\Size $ k $} & \makecell[c]{Dilation \\Rate $ d $} & mAP(\%) \\
    			\hline
    			\cite{Yeung2018} Two-stream     &     -     &     -     & 27.60 \\
    			\cite{Yeung2018} Two-stream + LSTM     &     -     &     -     & 28.10 \\
    			\cite{Yeung2018} Multi-LSTM    &     -     &     -     & 29.60 \\
    			\cite{DBLP:conf/cvpr/DaveRR17} Predictive-corrective     &     -     &     -     & 29.70 \\
    			\cite{ZhaoXWWTL17} SSN    &     -     &     -     & 30.30 \\
    			\cite{DBLP:conf/cvpr/PiergiovanniR18} I3D + super-events     &     -     &     -     & 36.40 \\
    			\cite{DBLP:conf/icml/PiergiovanniR19} I3D + super-events + TGMs    &     -     &     -     & 46.40 \\
    			\cite{Hussein2019} I3D  &     -     &     -     & 72.43 \\
    			\cite{Hussein2019} I3D + Timeception &     3     &     1     & 72.83 \\
    			\cite{Hussein2019} I3D + Timeception &     3     &    1,2,3    & 74.52 \\
    			\cite{Hussein2019} I3D + Timeception &    1,3,5,7    &     1     & 74.79 \\
    			\hline
    			I3D + NAS-TC  &    3,5,7    &     1,2     & \textbf{76.83} \\
    			\hline
    		\end{tabular}
    			
    		\label{tab:MultiTHUMOS}
    	\end{center}
    	\vspace{-0.2cm}
    \end{table}

	\subsection{Experiments on BreakfastAction}
	We also test the performance of our model on the BreakfastAction dataset. For ease of comparison, we use the same settings as in \cite{Hussein2019} of which uniformly sample 64 segments from each video, each of which contains eight successive frames, and then use a model of three stacked layers of NAS-TC layers to model the time dimension. The performance comparison of the models is presented in Table \ref{tab:BreakfastAction}. It can be seen from the experimental results that the performance of NAS-TC still maintains corresponding advantages.

	\begin{table}[htbp]
		\begin{center}
		\caption{Performance comparison between NAS-TC and other related methods on the single-label classification of activities and multi-label classification of unit-actions on the BreakfastAction dataset.}
			\begin{tabular}{lcc}
				\hline
				Method   & \makecell[c]{Activities \\(Acc. \%)} & \makecell[c]{Actions \\(mAP \%)}\\
				\hline
				\cite{sigurdsson2017asynchronous} ResNet-152    &    41.13     &     32.65 \\
				\cite{girdhar2017actionvlad} ResNet-152 + ActionVLAD    &    55.49     &     47.12 \\
				\cite{Hussein2019} ResNet-152 + Timeception &    57.75     &     48.47 \\
				\cite{hussein2019videograph} ResNet-152 + VideoGraph    &     59.12  & 49.38 \\
				\cite{Carreira2017} I3D    &   58.61     &     47.05 \\
				\cite{girdhar2017actionvlad} I3D + ActionVLAD    &   65.48     &     60.20 \\
				\cite{Hussein2019} I3D + Timeception &    67.07     &     61.82     \\
				\cite{hussein2019videograph} I3D + VideoGraph &    69.45     &     63.14     \\ \hline
				I3D + NAS-TC &     \textbf{71.89}     & \textbf{64.64}   \\ 
				\hline
			\end{tabular}
			
			\label{tab:BreakfastAction}
		\end{center}
		\vspace{-0.2cm}
	\end{table}	
	
	\section{Conclusion}
	\label{Sec: Conclusion}
	In this paper, a Neural Architecture Search Temporal Convolutional (NAS-TC) layer on the temporal dimension is proposed. We design a special search space for NAS, extend DARTS to Temp-DARTS, and design a two-phase framework for NAS applications in the long video context. In NAS-Temp, the repeated use of intermediate nodes gives the model sufficient ability to learn long-range temporal dependencies. In addition, the parameter assignment of the model follows the principle of subspace efficiency effectively. This allows NAS-TC to still achieve good performance using relatively few parameters. NAS-TC also maintains the flexible automation advantages of NAS. Our method is also very scalable and provides a good reference for solving similar problems in other fields.

\section*{Acknowledgment}
This work was partially supported by the NSFC under Grant (No.61972315 and No. 62072372), the Shaanxi Science and Technology Innovation Team Support Project under grant agreement (No.2018TD-026) and the Australian Research Council Discovery Early Career Researcher Award (No.DE190100626).

\ifCLASSOPTIONcaptionsoff
  \newpage
\fi



%
\bibliographystyle{IEEEtran}
\bibliography{BibTex}




%
%
%





\end{document}